\documentclass{article} 
\usepackage[table]{xcolor}
\usepackage{iclr2024_conference,times}

\usepackage{amsmath,amsfonts,bm}

\def\eqref#1{equation~\ref{#1}}

\def\1{\bm{1}}

\DeclareMathAlphabet{\mathsfit}{\encodingdefault}{\sfdefault}{m}{sl}
\SetMathAlphabet{\mathsfit}{bold}{\encodingdefault}{\sfdefault}{bx}{n}

\usepackage{textcomp}
\usepackage{diagbox}
\usepackage{hyperref}       
\usepackage{url}            
\usepackage{booktabs}       
\usepackage{amsfonts}       
\usepackage{nicefrac}       
\usepackage{microtype}      
\usepackage{bbm}          
\usepackage{xcolor}         
\usepackage{booktabs}
\usepackage{graphicx}
\usepackage{multirow}
\usepackage{wrapfig}
\usepackage{float}
\usepackage{bbm}
\usepackage{listings}
\usepackage{fancyhdr}
\usepackage{enumitem}
\usepackage{cases}
\usepackage{xcolor}
\usepackage{ulem}
\setlist[itemize]{leftmargin=*}
\usepackage[ruled,vlined]{algorithm2e}
\definecolor{commentcolor}{RGB}{110,154,155}

\definecolor{pearThree}{HTML}{E74C3C}
\definecolor{pearDarker}{HTML}{1D2DEC}
\definecolor{darkTurquoise}{HTML}{007C7D}
\hypersetup{
	colorlinks,
	citecolor=darkTurquoise,
	linkcolor=pearThree,
	breaklinks=true,
	urlcolor=pearDarker}
\definecolor{shadecolor}{rgb}{0.92,0.92,0.92}

\title{MarineGPT:  Unlocking Secrets of ``Ocean'' to the Public}

\author{Ziqiang Zheng\textsuperscript{1}, Jipeng Zhang\textsuperscript{1}, Tuan-Anh Vu\textsuperscript{1}, Shizhe Diao\textsuperscript{1}, Yue Him Wong Tim\textsuperscript{2}, Sai-Kit Yeung\textsuperscript{1} \\
\textsuperscript{1}Hong Kong University of Science and Technology,  \textsuperscript{2}Shenzhen University\\
\small{\texttt{\{zzhengaw,jzhanggr,tavu,sdiaoaa\}@connect.ust.hk}, \texttt{saikit@ust.hk}} \\
}

\iclrfinalcopy 

\begin{document}
\maketitle

\begin{abstract}
Large language models (LLMs), such as ChatGPT/GPT-4, have proven to be powerful tools in promoting the user experience as an AI assistant. The continuous works are proposing multi-modal large language models (MLLM), empowering LLMs with the ability to sense multiple modality inputs through constructing a joint semantic space (\emph{e.g.} visual-text space). Though significant success was achieved in LLMs and MLLMs, exploring LLMs and MLLMs in domain-specific applications that required domain-specific knowledge and expertise has been less conducted, especially for \textbf{marine domain}. Different from general-purpose MLLMs, the marine-specific MLLM is required to yield much more \textbf{sensitive}, \textbf{informative}, and \textbf{scientific} responses. In this work, we demonstrate that the existing MLLMs optimized on huge amounts of readily available general-purpose training data show a minimal ability to understand domain-specific intents and then generate informative and satisfactory responses. To address these issues, we propose \textbf{MarineGPT}, the first vision-language model specially designed for the marine domain, unlocking the secrets of the ocean to the public. We present our \textbf{Marine-5M} dataset with more than 5 million marine image-text pairs to inject domain-specific marine knowledge into our model and achieve better marine vision and language alignment. Our MarineGPT not only pushes the boundaries of marine understanding to the general public but also offers a standard protocol for adapting a general-purpose assistant to downstream domain-specific experts. We pave the way for a wide range of marine applications while setting valuable data and pre-trained models for future research in both academic and industrial communities. Code and data will be available at \textcolor{blue}{\href{https://github.com/hkust-vgd/MarineGPT}{https://github.com/hkust-vgd/MarineGPT}}
\end{abstract}

\section{Introduction}
Large language models (LLMs)~\citep{ouyang2022training,openai2023gpt4} demonstrated an impressive ability for a large range of user-tailored tasks, endowing them with immense potential applications~\citep{fu2023mme,cox2023utilizing,tu2023towards,hong20233d}. As a general-purpose assistant, ChatGPT/GPT-4~\citep{openai2023gpt4,ouyang2022training} could understand human intents and complete various real-world tasks. However, existing LLMs only focus on unimodal text inputs. Obviously, a text-only Chatbot is less optimal for a powerful AI assistant. The multi-modal large language models (MLLMs)~\citep{li2023videochat,liu2023visual,zhao2023chatbridge,zang2023contextual,wu2023visual} empower the LLMs to sense multiple modality inputs, follow multi-modal instructions, and align with human intent to complete various real-world tasks in the wild. A vision-language multi-modal model~\citep{zhu2023minigpt} connects a vision encoder and LLM~\citep{touvron2023llama} for general-purpose visual-and-language understanding. There has been rapid progress in the MLLM field by leveraging billions of image-text pairs~\citep{kakaobrain2022coyo-700m,schuhmann2022laion} from the public web. Vision-language pre-training~\citep{li2022blip,li2023blip} is conducted for bridging the frozen pre-trained vision encoders and frozen language decoders. However, such general-domain vision-language models still lack sophistication and conception coverage in understanding and delivering domain-specific knowledge. MiniGPT-4~\citep{zhu2023minigpt} proposed to align a frozen visual encoder with a frozen LLM by optimizing the linear layer to bridge the two modalities. LLaVa~\citep{liu2023visual} conducted the first attempt to perform the visual instruction tuning, enabling the LLMs to understand the user intent and generate corresponding responses.

\begin{figure}[t]
\begin{center}
\includegraphics[width=\textwidth]{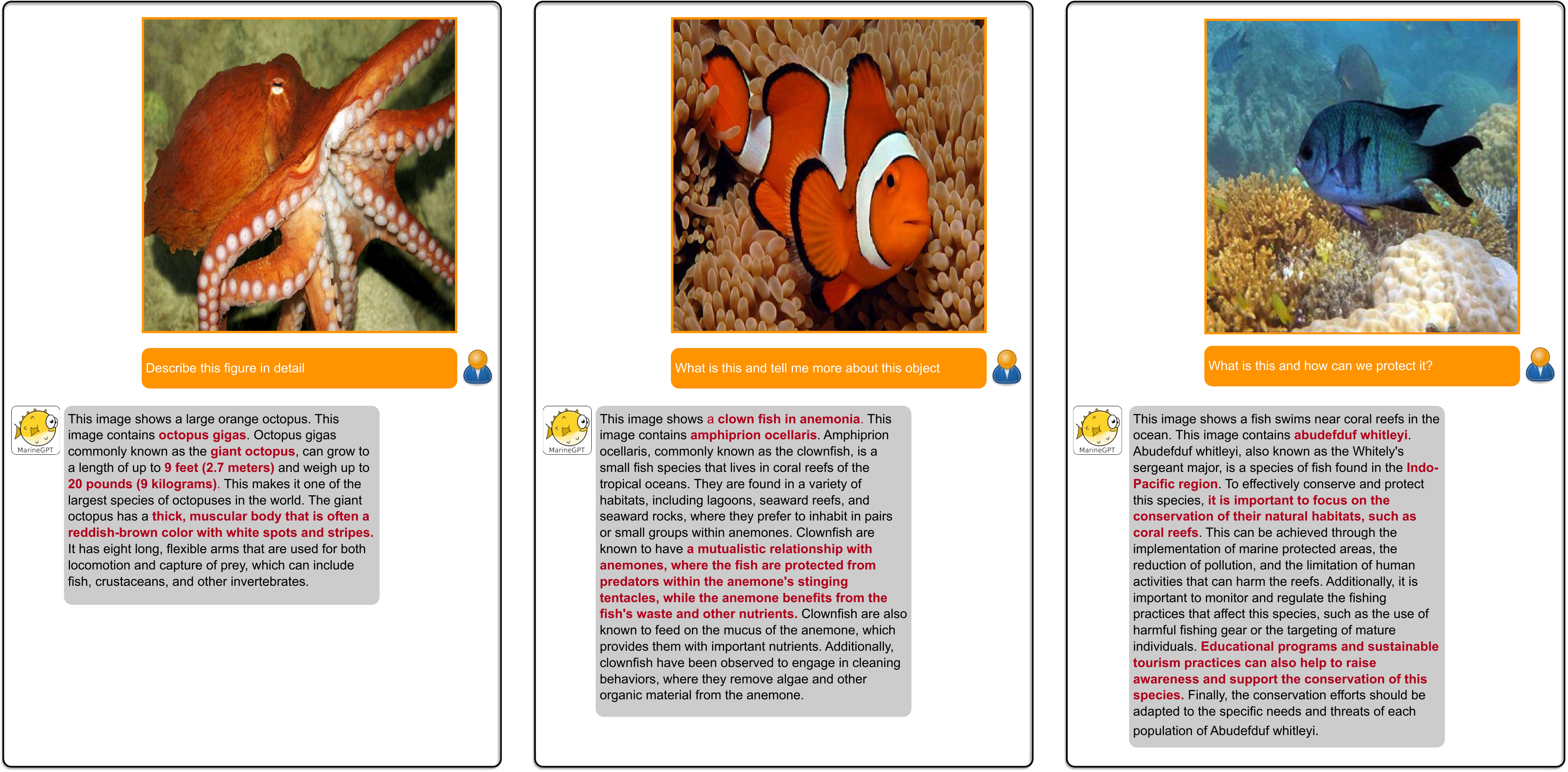}
\end{center}
\caption{MarineGPT could perform auto-recognition of various marine objects and yield \textit{diverse}, \textit{domain-specific}, \textit{informative}, and \textit{scientific} responses associated with the recognized marine object. Best viewed in color.}
\label{fig:demo}
\end{figure}

To enable MLLMs to understand the user intent and generate domain-specific answers, further efforts and elaborate designs are specially required. The dominant way is to collect moderate-scale but high-quality image-text pairs from the specific domains for downstream fine-tuning~\citep{zhang2023pmc}. LLaVa-Med~\citep{li2023llava} proposed to fine-tune the LLAVa~\citep{liu2023visual} to the medicine domain to generate the domain-specific responses. The instruction-following training samples have also been meticulously curated for the medical domain to enable the model to understand the user intent from the specific domains. In this work, we propose \textbf{MarineGPT} presented in Fig.~\ref{fig:demo}, the first marine-specific vision-to-language expert model, which could recognize marine objects from the given visual inputs and yield corresponding \textit{sensitive}, \textit{informative}, and \textit{scientific} responses as a powerful marine AI assistant. Developing an effective marine AI assistant poses three non-trivial challenges: 1) the existing general-purpose LLMs are \textit{less-efficient} to generate reasonable and scientific responses for the marine domain; 2) \textit{limited concept coverage} for the marine domain present in existing image-text pairs; and 3) limited ability to perform the \textit{fine-grained object recognition} since many marine species share very similar appearance. 

To address these challenges, we first scale up the marine knowledge collection through crowdsourcing multiple marine sources for collecting huge amounts of training data. We present our \textbf{Marine-5M} dataset that contains more than 5 million marine image-text pairs with \textit{redundant conceptions from the marine domain}. We utilize our Marine-5M for \textbf{marine-specific continuous pre-training}, which could effectively adapt the general-purpose MLLM to the domain-specific expert model, aligning images with domain expertise flexibly defined and managed based on the language descriptions. We then design \textbf{50} different \textbf{marine-specific instructions} based on the expertise and requirements from marine biologists, which could help MarineGPT understand the user intent. We scalably generate instruction-following training data based on ChatGPT/GPT-4, following our constructed marine-specific instructions. Besides, we also summarized \textbf{129} diverse and comprehensive attributes (\emph{e.g.} distribution, habitat, morphology, reproduction, and \textit{etc}) of marine objects. We retrieved and collected corresponding descriptions based on summarized attributes and crawled category annotation from reliable marine knowledge sources~\citep{eol,yang2020fishdb,corals,reeflex}. Through these ways, we can effectively inject marine knowledge into the model and enable MarineGPT to generate informative and domain-specific responses. The constructed instruction-following training data (with 1.12 million high-quality marine image-text pairs) are utilized for \textbf{instruction-following fine-tuning} demonstrated in Fig.~\ref{fig:framework}. Finally, we empirically demonstrate that only designing the linear layer~\citep{zhu2023minigpt} cannot effectively align the visual signals and the textual descriptions for fine-grained marine object recognition. The Q-Former should also be optimized for more effective and fine-grained marine vision-language alignment. 

MarineGPT is unlocking the secrets of the ocean to the public. It provides an effective way to perform non-invasive automated species recognition, reducing human labor from the domain experts. We observe that MarineGPT could effectively identify marine organisms even under a fine-grained setting, enabling scientific image annotation and decision-making processes. Meanwhile, MarineGPT could also enable complex visual reasoning, knowledge-grounded image description, and multi-turn conversations, allowing citizens, scientists, and the general public to participate actively in marine research and conservation. In summary, our main contributions are listed as follows.
\begin{itemize}
    \item MarineGPT empowers marine object recognition with question-answering capabilities, which could encourage public participation in marine research and data collection. Both citizens and scientists could contribute to biodiversity monitoring and research efforts, helping to gather valuable information for marine studies.
    \item We propose the largest marine image-text pairs as far as we know for promoting aligning visual-and-language modalities and enhancing the model's perception capabilities. The collected and constructed diverse and comprehensive image-text pairs could significantly change the way of marine perception, reasoning, and causal inference.
    \item We present a marine-specific data generation pipeline to create diverse (image, instruction, output) training data, by sampling marine image-text pairs from broad-coverage and open-source websites and using ChatGPT/ChatGPT-4 to create instruction-following data. 
    \item We provide a unified and valuable perspective for performing domain-specific cross-modal vision-language tasks, which require professional knowledge injection and user intent understanding. Similar approaches could be adopted in other fields like botany, entomology, or ornithology, enabling species recognition and comprehensive data analysis.
\end{itemize}

\section{Related Work}
\subsection{LLMs}
LLMs have demonstrated remarkable progress across various Natural Language Processing (NLP) tasks. The impressive performance of ChatGPT~\citep{chatgpt} and GPT-4~\citep{openai2023gpt4} has motivated the research community to reproduce their success, resulting in a rapidly increasing number of foundation models. 
These efforts include GPT-3~\citep{brown2020gpt3}, T5~\citep{raffel2020exploring}, BLOOM~\citep{scao2022bloom}, OPT~\citep{zhang2022opt}, LLaMA~\citep{touvron2023llama}, and LLaMA-2~\citep{touvron2023llama2}, all of which have get worldwide significant interest. 

\noindent\textbf{Pre-training LLMs for domain-specific applications}. In addition to general-purpose LLMs, several LLMs are specifically pre-trained from scratch to cater to domain-specific tasks. These domain-specific pre-trained LLMs, such as Galactica~\citep{taylor2022galactica} and CodeGen~\citep{nijkamp2022codegen}, are designed to address specific domains like science and program synthesis~\citep{lu2022learn}. 

\noindent\textbf{Fine-tuning LLMs for domain-specific applications}. These foundation models have resulted in the exponential growth of fine-tuned variants in task-oriented settings, such as chatbots. In order to enable these LLMs to follow human commands and complete tasks, instruction tuning is introduced to obtain these models' instruction-following versions like InstructGPT~\citep{ouyang2022training}, Flan-T5~\citep{chung2022scaling}, and Vicuna~\citep{chiang2023vicuna}. Additionally, various domain-specific applications have emerged based on these foundation models, including Dr.LLaMA~\citep{guo2023drllama} and CancerGPT~\citep{li2023cancergpt} for medical applications, as well as FinGPT~\citep{yang2023fingpt} and BloombergGPT~\citep{wu2023bloomberggpt} for finance. Also, several toolboxes for fine-tuning LLM, like LMFlow~\citep{diao2023lmflow}, have been developed. The explosive emergence of those task-oriented LLMs leads to impressive performance gains and opens up new possibilities for various LLM applications.

\subsection{MLLMs}
As LLMs grow more intelligent and powerful, there is increasing attention on extending the frozen LLMs to multi-modal tasks, empowering the LLMs with the ability to sense multi-modality inputs. Flamingo~\citep{alayrac2022flamingo} pioneered to harness web-scale image-text data, capitalizing on both vision and language models. This unveiling showcased remarkable zero-shot image-text capabilities in a conversational format. BLIP~\citep{li2022blip,li2023blip} bootstraps vision-language pre-training from frozen pre-trained image encoders and frozen language decoders. Based on BLIP-2~\citep{li2023blip}, MiniGPT-4~\citep{zhu2023minigpt} proposed a projection layer to align pre-trained vision encoder to frozen LLMs (\emph{e.g.} Vicuna~\citep{chiang2023vicuna}), and exhibited respectable zero-shot image comprehension in dialogues. Since only linear layers are required to be optimized, MiniGPT-4 offers high computation efficiency and flexibility for being fine-tuned to different downstream tasks. However, we noticed that only optimizing the linear layers cannot effectively adapt the general-purpose MLLM to a domain-specific expert model.  

\noindent\textbf{MLLMs for domain-specific applications}. A growing body of research has emerged to extend general-purpose MLLMs to domain-specific applications. LLaVa-Med~\citep{li2023llava} proposed to fine-tune the LLAVa to the medicine domain. Pathasst~\citep{sun2023pathasst} collected over 142K high-quality pathology image-text pairs from various reliable sources. Pmc-vqa~\citep{zhang2023pmc} proposed to sample biomedical image-text pairs from PMC-15M and adopt GPT-4 to create instructions from the text alone. EgoCOT~\citep{mu2023embodiedgpt} craft a large-scale embodied planning dataset, consisting of carefully selected videos from the Ego4D dataset for effective embodied planning. In this work, we aim to propose the first marine-specific MLLM MarineGPT, and a large collection of high-quality marine image-text pairs, which covers a broad spectrum of marine and marine-relative conceptions. 

\section{Approaches}
\subsection{Framework Overview}
\begin{figure}[t]
\begin{center}
\includegraphics[width=0.9\textwidth]{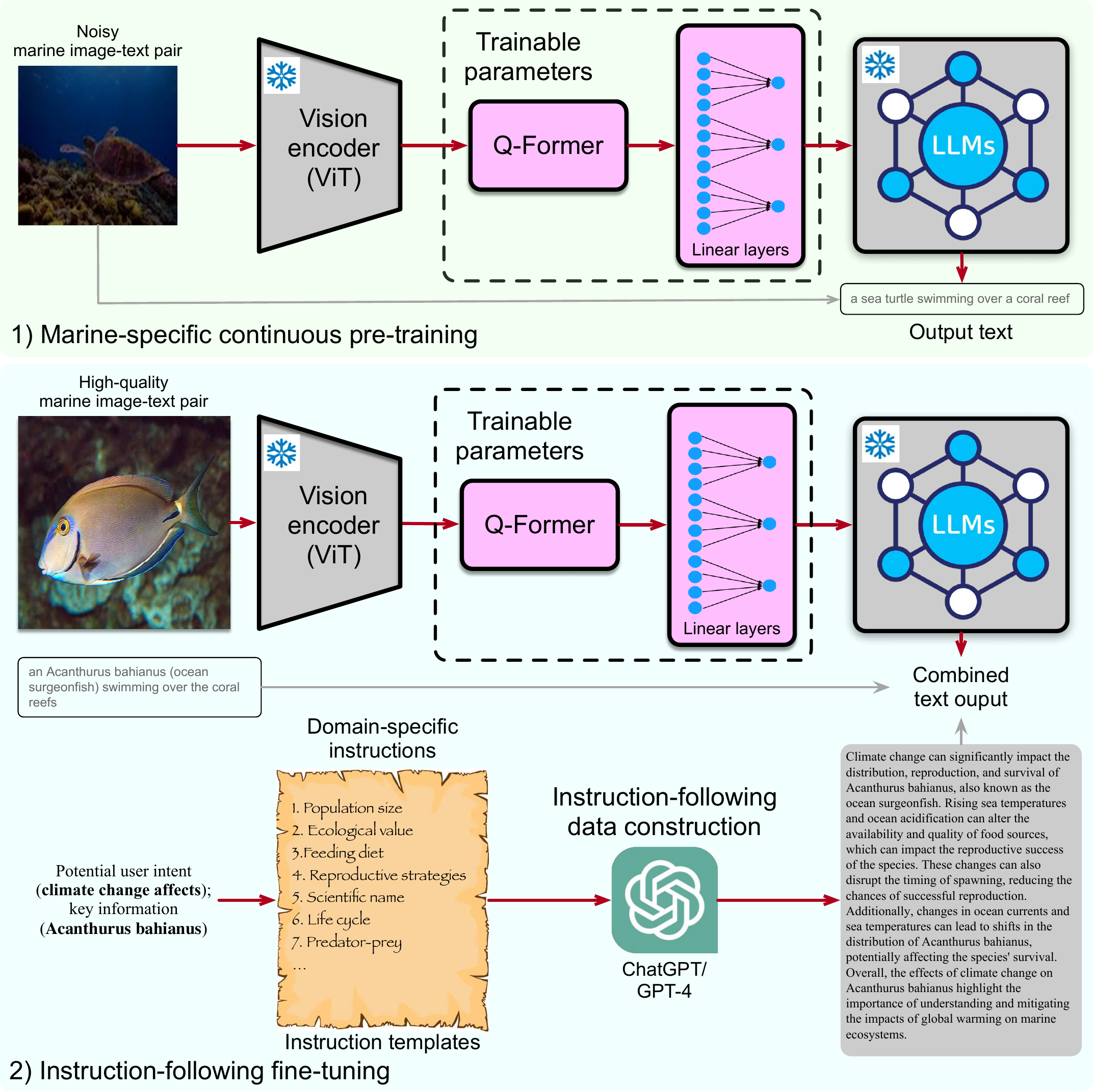}
\end{center}
\caption{The framework overview of the proposed MarineGPT. There are two main procedures in our MarineGPT: 1) marine-specific continuous pre-training on 5 million marine image-text pairs; 2) instruction-following fine-tuning based on constructed high-quality instruction-following image-text pairs to generate \textit{sensitive}, \textit{informative} and \textit{scientific} responses.}
\label{fig:framework}
\end{figure}

The framework overview of MarineGPT is demonstrated in Fig.~\ref{fig:framework}. To better utilize the powerful zero-shot ability of LLMs and adapt them to domain-specific experts, we design a two-stage training approach: 1) a marine-specific continuous pre-training on a large collection of aligned marine image-text pairs for acquiring marine vision-language alignment; 2) a further instruction-following fine-tuning based on a moderate scale but high-quality self-constructed image-text pairs with specifically designed instructions to generate more informative, reliable and scientific answers, enhancing the professional ability and usability of MarineGPT. The former continuous pre-training procedure could effectively inject marine knowledge into the LLM. Meanwhile, broad coverage of marine conceptions that have been overlooked in existing general-purpose LLMs and MLLMs, could be introduced and emphasized through continuous pre-training. Moreover, the constructed instruction-following data could promote generating domain-specific responses with the required information delivered and a better understanding of the user intent.

\subsection{Marine-Specific Continuous Pre-training}
The vision-language pre-training based on redundant image-text pairs is an \textit{essential}, \textit{important}, and \textit{foundational} step to acquiring vision-language knowledge and constructing a joint feature space. To scale up data collection, the readily available image-text pairs from open-source websites are used for vision-language pre-training. We adopt the pre-trained ViT encoder of BLIP-2~\citep{li2023blip} optimized by CC12M~\citep{changpinyo2021conceptual}, COCO captions~\citep{lin2014microsoft}, Visual Genone~\citep{krishna2017visual} and sub-part of LAION dataset~\citep{schuhmann2021laion,schuhmann2022laion}. In total, there are 14 million image-text pairs during the initial pre-training stage in ~\citep{li2022blip}. The existing general-domain LLMs are less effective for marine scenarios because marine image-text pairs drastically differ from general web content. The current MLLMs cannot generate or answer the required or accurate descriptions based on the given image. There are two main drawbacks when directly utilizing the pre-trained models for instruction-following fine-tuning: 1) the existing large collection of aligned image-text pairs does \textbf{not} involve too much marine knowledge or marine visual observations and the conception coverage is not enough to include most marine organisms; 2) the description of images are usually short (especially for the LAION dataset~\citep{schuhmann2021laion,schuhmann2022laion}), which cannot reveal and deliver too much detail information. Thus, the pre-trained VIT encoder would still be unable to extract effective feature representations from the marine images. Meanwhile, the language decoder will also fail to generate long and meaningful descriptions. The existing general-purpose LLMs are less effective for marine scenarios because marine image-text pairs drastically differ from general web content. To address these two issues, we present \textbf{Marine-5M} dataset with 5 million marine image-text pairs to promote the ability to extract reliable marine features. Our constructed Marine-5M is the largest marine image-text dataset specially designed for marine research. 

Different from the pre-training stage of BLIP-2~\citep{li2023blip}, we optimize the Q-Former and the linear layer architectures to bootstrap the marine vision-language pre-training based on our  Marine-5M dataset. Considering the annotation of crawled images from public websites only provides the category annotation, we propose to expand the caption of images according to \textbf{$<$category annotation$>$}. We summarized \textbf{129} diverse, comprehensive, and hierarchical attributes of the marine objects, including size, color, shape, feeding diet, distribution, habitat, morphology, reproduction, and \textit{etc}. We generate different attribute descriptions based on texts crawled from reliable marine websites (\emph{e.g.}, EOL~\citep{eol}, FishDB~\citep{yang2020fishdb}, Reeflex~\citep{reeflex} and so on) for images with the same category annotation. In this way, we could generate longer image descriptions and perform the marine knowledge injection. After marine-specific continuous pre-training, the trained model demonstrates the capacity to possess a wealth of knowledge in the marine field and offer more accurate and reasonable responses to human inquiries. To further promote the ability of MarineGPT to generate more fine-grained and scientific responses, we construct \textbf{1.12 million} high-quality image-text pairs with a large range of instruction-following templates, presenting various tasks of describing marine organisms in the given image.

\subsection{Instruction-Following Fine-tuning}
Performing continuous pre-training on the marine image-text pairs alone is \textbf{insufficient} to build a domain-specific and scientific Chatbot for answering diverse marine questions, as it lacks the ability to follow diverse marine-specific instructions even though marine concept coverage is highly promoted after the continuous pre-training. Terribly, the existing LLMs would refrain from answering biology and marine questions, or worse, produce \textbf{incorrect responses} or \textbf{complete hallucinations}. In most scenarios, the existing general-purpose visual assistants may behave like a layperson for domain-specific questions and cannot understand the user intent. A simple and effective method of aligning LLMs to human intent is to learn from instruction-following data generated by state-of-the-art LLMs. 

\subsubsection{Instruction-Following Image-Text Data Construction}
MarineGPT adopts a wide range of vision-language instruction data, covering both template-based converted image captions and ChatGPT-generated question-answer data following the user intent demonstrated in Fig.~\ref{fig:framework}. Asking insightful marine questions is crucial for acquiring domain-specific knowledge and expanding the understanding of the marine world. We propose to prompt MarineGPT to extend the generation based on different user intent and domain-specific questions about the identified marine organism, such as its habitat, behavior, ecological role, conservation status, and \textit{etc.} 

\noindent\textbf{Instruction-following data construction}. We generate diverse and comprehensive multi-modal vision-language instruction-following data. We construct \textbf{50} different instructions and scalably invoke instruction-following data based on crawled marine knowledge from both open-source public \textbf{marine websites} and \textbf{ChatGPT/GPT-4}. We provide some designed instructions as follows:
\begin{itemize}
    \item \textbf{Instruction 1.} ``\textit{please describe the species \textbf{richness} and \textbf{distribution} of \textbf{$<$image category$>$}}.''
    \item \textbf{Instruction 2.} ``\textit{please answer what are the \textbf{predator-prey relationships} for the \textbf{$<$image category$>$} and how they influence population dynamics}.''
    \item \textbf{Instruction 3.} ``\textit{please answer how this \textbf{$<$image category$>$} \textbf{interacts with other species} in marine ecosystems}.''
    \item \textbf{Instruction 4.} ``\textit{please answer the \textbf{conservation status} of \textbf{$<$image category$>$}, including their \textbf{population trends}, \textbf{threats} they face, and the \textbf{effectiveness of existing conservation measures}}.''
\end{itemize}
The constructed instruction-following data will be utilized for optimizing our MarineGPT to generate more user-centric and scientific responses. 

\begin{figure}[t]
\begin{center}
\includegraphics[width=\textwidth]{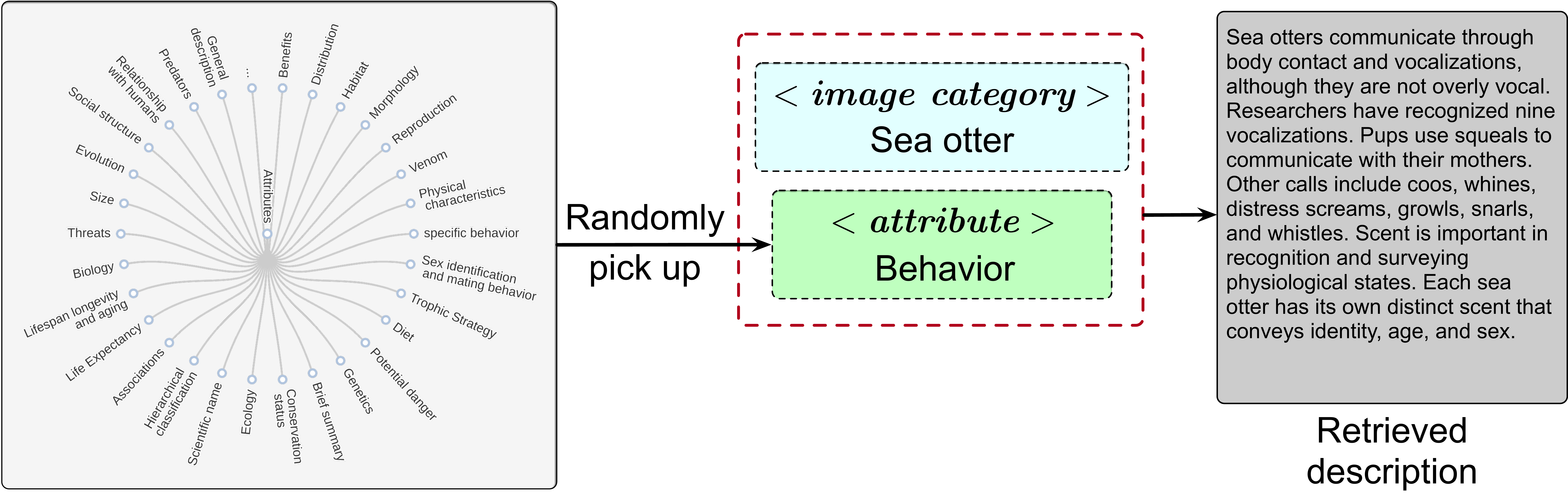}
\end{center}
\caption{The procedure of our attribute-based image description generation pipeline.}
\label{fig:attribute}
\end{figure}

\noindent\textbf{High-quality image-text pair formulation}. Besides leveraging the capabilities of ChatGPT/GPT-4 to formulate the instruction-following training data. We also follow our summarized attribute-based image description generation pipeline (refer to Fig.~\ref{fig:attribute}). By combining web-scraped image-text data from open-source marine websites and domain-specific knowledge, we could enable MarineGPT to generate more \textit{reliable}, \textit{accurate}, and \textit{scientific} answers aligned with the user intent. We dedicate substantial efforts to gathering data from diverse sources, including books, biology blogs, scientific articles, and open-source marine websites. The constructed image-text pairs from a broad spectrum of sources, including public websites, various open-source books, guidebooks, and our privately collected diving and surveying data. After sophisticated and extensive data cleaning and caption alignment, we generate longer and more domain-specific image captions.

We generate \textbf{1.12 million} scientifically constructed instruction-following question-answer pairs for marine knowledge foundation model training. This marine-specific data generation approach enables us to generate more image-text pairs with detailed and informative image descriptions. To facilitate marine multi-modal research, we will release our constructed marine-specific instruction-following data and the trained MarineGPT model. The marine and biology researchers could then further formulate their own high-quality vision-language datasets, thereby promoting advancements in the marine foundation model community.

\subsubsection{Instruction-Following Fine-tuning}
MarineGPT could produce more detailed and scientific image descriptions with much more diversity and align with the user intent. We provide some prompt templates as shown below:
\begin{itemize}
    \item \textbf{Prompt 1.} ``\textit{Can you answer what \textbf{ecological} roles does the marine organism in the \textbf{$<$image$>$} play in their \textbf{ecosystems}}?''
    \item \textbf{Prompt 2.} ``\textit{Could you describe how do \textbf{climate changes} affect the \textbf{distribution}, \textbf{reproduction}, and \textbf{survival} of the object in the \textbf{$<$image$>$}}?''
    \item \textbf{Prompt 3.} ``\textit{Please determine the \textbf{scientific name} of object in this \textbf{$<$image$>$}, classification within the \textbf{taxonomic hierarchy}, and its \textbf{relationships} to \textbf{other known species}}.''
    \item \textbf{Prompt 4.} ``\textit{Take a look at this image and describe how can we \textbf{mitigate} and \textbf{} \textbf{human-induced threats} to the object in this \textbf{$<$image$>$}}.''
\end{itemize}
Please note that the image prompts are aligned with the previous 50 constructed instructions. We conduct instruction-following fine-tuning to the marine domain for end-to-end training of a marine visual-to-language conversational assistant, which could promote the capabilities and usability of AI-powered systems, making them more versatile, context-aware, and user-centric. 
\section{Experiments}

\subsection{Implementation Details}
MarineGPT aims to achieve cross-modality visual-and-language alignment between the visual observations and LLMs. To achieve effective visual perception, we adopt the same visual encoder as used in BLIP-2~\citep{li2023blip}, a ViT backbone with a pre-trained Q-Former. For the large language model, we adopt LLaMA-13B~\citep{touvron2023llama} as the decoder to generate responses. Both language and vision models are open-sourced. We target to bridge the gap between the visual encoder and LLM based on the Q-Former along with additional linear layers for computing the similarity between the visual content and captions. Through such pre-training, we could perform the marine vision-language alignment.

Additionally, we convert the data type of parameters of the frozen ViT~\citep{dosovitskiy2020image} and language model to FP16 during the continuous pre-training to increase computational efficiency. At the first continuous marine-specific pre-training, we focus on marine image-text alignment, which involves using our constructed Marine-5M dataset with a large range of marine image-text pairs. To relieve the computation burden brought by large-scale image-text pairs, we adopt a low-resolution input ($224\times 224$) for projecting image-text pairs into the same semantic space. At the second instruction-following fine-tuning procedure, we adopt a larger image resolution ($384\times 384$) to provide more detailed information. In the second fine-tuning procedure, we only adopt the constructed 1.12 million high-quality marine image-text pairs (with both the template-based image captions and our instruction-following ChatGPT/GPT-4 generated captions) to enable MarineGPT more powerful.

\subsection{Marine-5M Dataset}
\textbf{Dataset construction}. To promote the image diversity of our \textbf{Marine-5M} dataset, we include marine image-text pairs from multiple different sources, including YouTube, Flickr, Google image engine, open-source marine websites (\emph{e.g.}, FishDB~\citep{yang2020fishdb}, Fishes-of-Australia~\citep{australia}, Corals-of-World~\citep{corals}, Reeflex~\citep{reeflex}, \textit{etc}) and publicly available marine datasets. Our Marine-5M dataset contains about \textbf{100K} marine and marine-relative object conceptions. For video sequences from YouTube, we especially download marine documentary videos, which contain high-quality marine image-text pairs. We utilize the subscripts to clip the keyframes and generate the aligned marine image-text pairs. The human-based checking and refinement are conducted to get more accurate and reliable text descriptions. The diving, surveying, and transect videos are also included to promote the diversity of our Marine-5M dataset. We extract one keyframe every 10 seconds for those videos without subscripts. The images from our Marine-5M dataset have a large diversity, in which the \textit{appearance}, \textit{boundary}, \textit{shape}, and \textit{texture} of marine objects vary significantly. Some images are captured in the wild and with \textit{low visibility}, \textit{background clutter}, \textit{motion blur}, \textit{occlusion}, \textit{dynamic illumination}, \textit{color distortion}, and \textit{optical artifacts}. The \textbf{Marine-5M} dataset also contains web-scraped image-text data from open-source marine websites. These web-scraped image-text pairs usually contain the category annotation and some brief summary of the marine object present in the visual images. Such pairs are usually high-quality and provide clean object conceptions centralized in images, which could promote the marine vision-language alignment.

As for the text descriptions, we expand the text description based on the provided category annotation considering that the category annotation or the image caption is over-simple. We summarize \textbf{129 different attributes} from the category annotation as demonstrated in Fig.~\ref{fig:attribute}. We generate corresponding captions based on these diverse attributes and category annotations. It is worth noting that not all marine object conceptions contain all these attribute annotations. The images with the same category annotation are assigned different attribute descriptions to promote the diversity of text captions and inject marine knowledge. For those images without any annotation, we propose to borrow the BLIP-2~\citep{li2023blip} to generate longer image captions for the marine images. We generate \textbf{5} diversified image caption candidates through Gaussian sampling in the latent space and then we rank all the generated image captions based on the caption length. Finally, we compute the sentence-level similarity between these caption candidates. Only the caption with a similarity less than \textbf{0.85} with the longest caption is appended for concatenation. In this manner, we could generate more diversified and longer captions for the marine images, describing more fine-grained color, texture, shape, and geometry information.

\begin{figure}[t]
\begin{center}
\includegraphics[width=\textwidth]{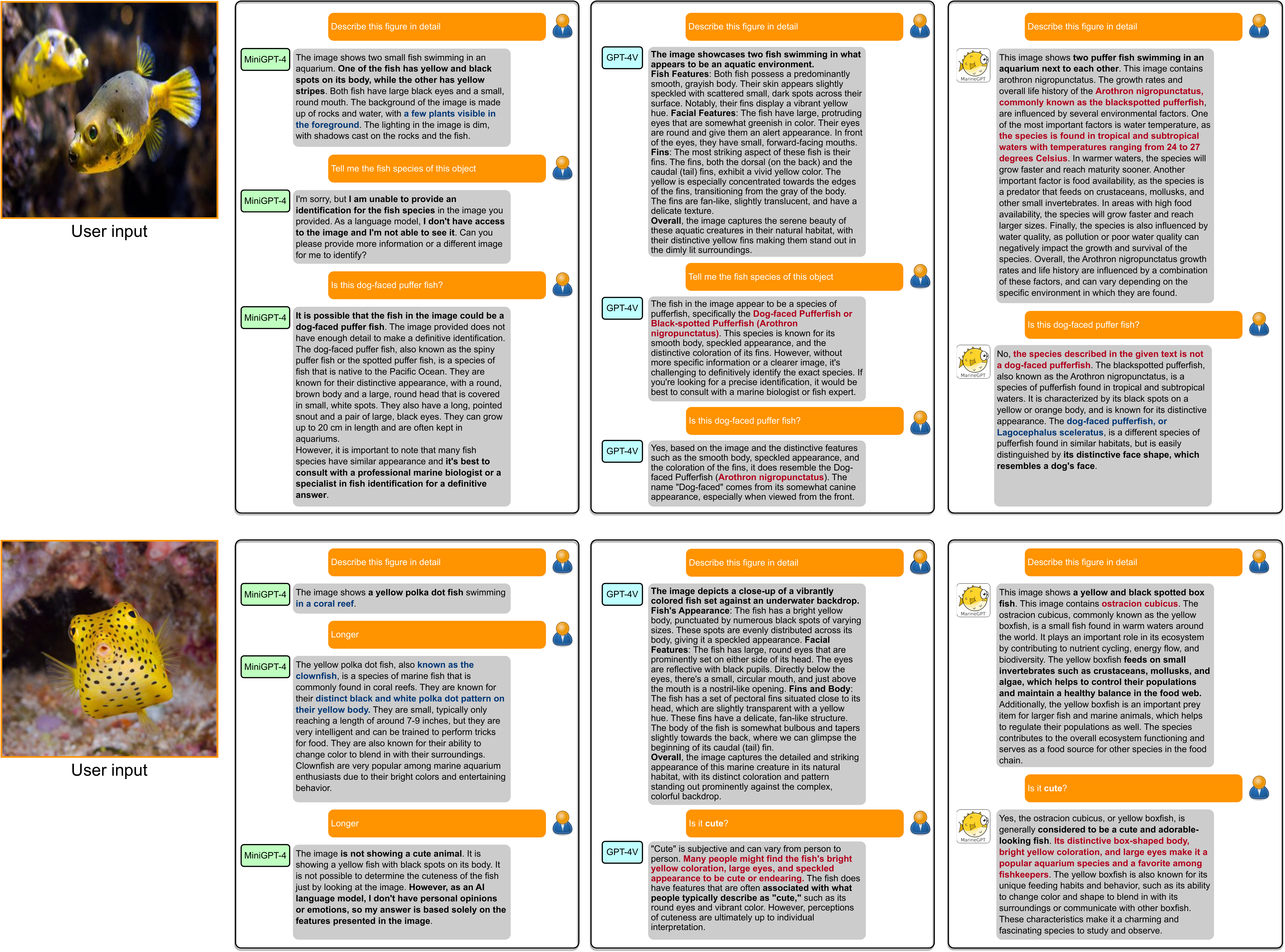}
\end{center}
\caption{The comparison between MiniGPT-4, GPT-4V and our MarineGPT. MarineGPT could recognize both the common and scientific names of marine objects and provide diverse information associated with the recognized objects. Best viewed in color.}
\label{fig:comparison}
\end{figure}

\subsection{Results}

\begin{figure}[t]
\begin{center}
\includegraphics[width=\textwidth]{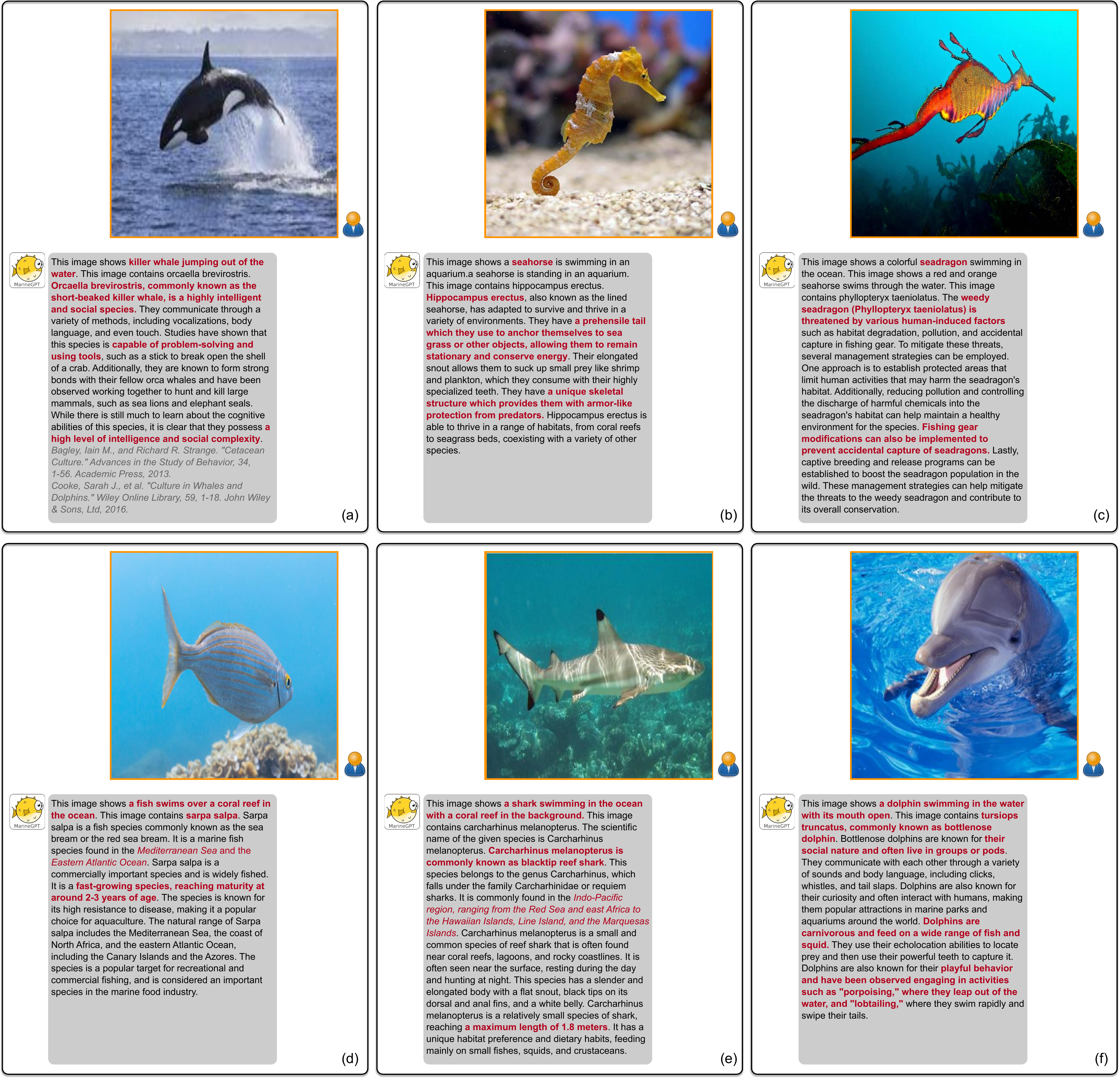}
\end{center}
\caption{MarineGPT could recognize a wide range of marine objects and yield comprehensive marine and biological knowledge delivered to the users so that the users could obtain a full understanding of the recognized marine objects.}
\label{fig:wide}
\end{figure}

\noindent\textbf{Comparison with MiniGPT-4 and GPT-4V}. We compare our MarineGPT with MiniGPT-4 and GPT-4V in Fig.~\ref{fig:comparison}. Compared with MiniGPT-4 and GPT-4V, our MarineGPT could generate long and detailed responses with corresponding biology information delivered (\emph{e.g.}, \textbf{scientific name} and \textbf{common name} of recognized objects). We observed that MiniGPT-4 will generate some descriptions (``\textit{a few plants visible in the foreground}'' and ``\textit{in a coral reef}'') that are \textbf{not} correct in the given marine images. We attribute this hallucination to the ineffective marine visual-language alignment. Directly optimizing the linear layer based on general-purpose image-text pairs will result in wrong descriptions for the marine images. GPT-4V cannot yield detailed marine-specific responses, either, while it tends to follow some fixed templates to describe the physical appearances of given images. In contrast, our MarineGPT could generate long and meaningful responses for domain-specific knowledge delivery. Both GPT-4V and MarineGPT could achieve accurate scientific name generation (``\textbf{Arothron nigropunctatus}'') and common name generation (``\textbf{Black-spotted Pufferfish}'') about the given marine object. However, MarineGPT is confused with dog-faced puffer fish (another common name of Arothron nigropunctatus) and generates a wrong scientific name for dog-faced puffer fish. In contrast, due to that GPT-4V could access the Internet, there is less possibility for GPT-4V to make such mistake. We attribute this failure of MarineGPT to the ineffectiveness of frozen LLM (LLaMA-13B~\citep{touvron2023llama} used in this work). It may still make mistakes for the domain-specific knowledge even if we performed the marine knowledge injection during the continuous vision-language pre-training. The further fine-tuning of LLMs to domain-specific domain may effectively alleviate such issues. Besides, Both GPT-4V and MarineGPT demonstrate \textbf{an aesthetic metric} summarized from common sense. Finally, compared with MiniGPT-4 and GPT-4V, MarineGPT could generate diverse and relative information associated with the recognized objects.

\noindent\textbf{Recognizing a wide range of marine objects}. We then provide more recognition results produced by our MarineGPT in Fig.~\ref{fig:wide}. MarineGPT could recognize a wide range of marine creatures and provide the corresponding common and scientific names of recognized objects. It is worth noting that our method could generate diverse and comprehensive image descriptions for the different recognized marine objects. MarineGPT could also generate the corresponding references to provide additional user information, as demonstrated in Fig.~\ref{fig:wide} a). It can generate informative responses describing how the recognized object benefits from the physical appearance shown in Fig.~\ref{fig:wide} b), how can we protect the recognized threatened marine species demonstrated in Fig.~\ref{fig:wide} c), the spatial distribution of ``sarpa salpa'' in Fig.~\ref{fig:wide} d), the physical characteristics of ``black-tip reef shark'' in Fig.~\ref{fig:wide} e) and the feeding diet and social behavior of ``bottlenose dolphin'' in Fig.~\ref{fig:wide} f). In this way, MarineGPT could deliver marine knowledge to the users.

\begin{figure}[t]
\begin{center}
\includegraphics[width=\textwidth]{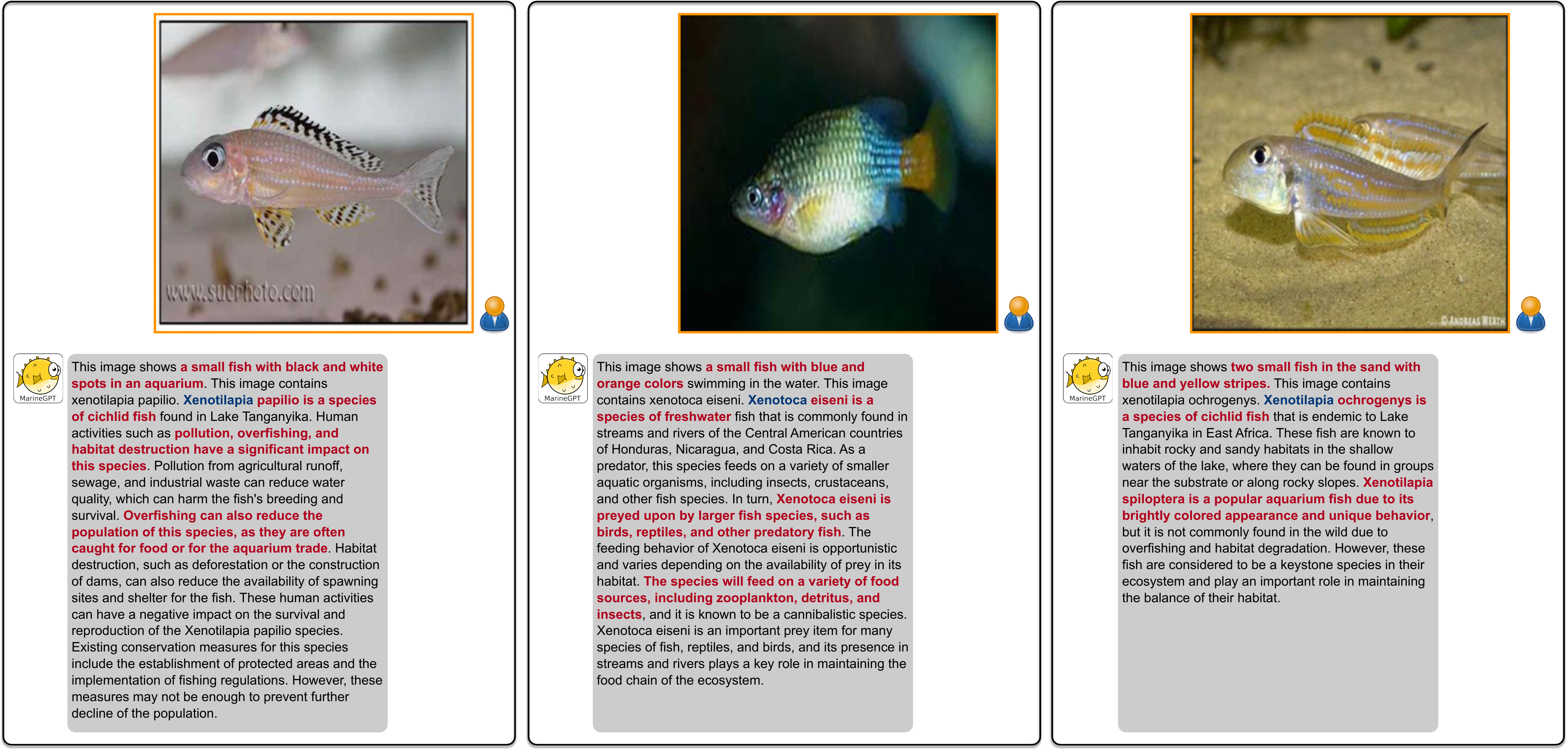}
\end{center}
\caption{Our MarineGPT could accurately recognize similar marine objects under the fine-grained setting.}
\label{fig:fine}
\end{figure}

\noindent\textbf{Fine-grained marine object recognition}. MarineGPT could discriminate very similar marine organisms and generate different responses based on them. We report the fine-grained object recognition results of MarineGPT in Fig.~\ref{fig:fine}. As demonstrated, MarineGPT could generate the corresponding scientific names for these three different fish species and provide further detailed descriptions from various aspects. This novel fine-grained object recognition ability of MarineGPT introduced by our Marine-5M dataset could enable diversity monitoring and reduce the human labor from the fish experts. The existing general-purpose MLLMs will regard all these three fish species as ``fish'' and fail to perform fine-grained recognition. 

\begin{figure}[t]
\begin{center}
\includegraphics[width=\textwidth]{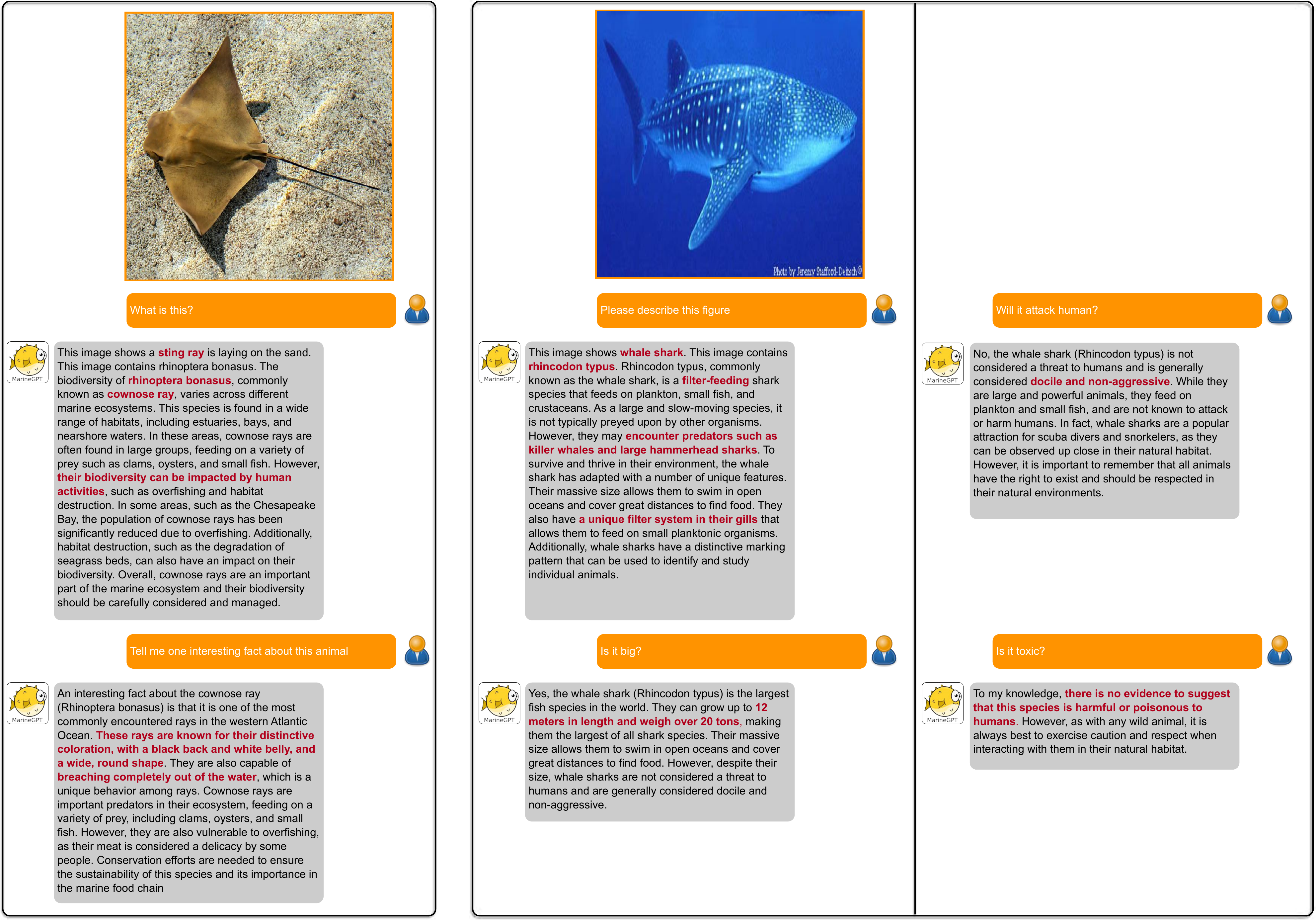}
\end{center}
\caption{Our MarineGPT could understand diverse the user intent and yield corresponding informative responses.}
\label{fig:conversation}
\end{figure}

\noindent\textbf{Multi-round conversation}. Users could upload images of different marine objects and ask different questions, as demonstrated in Fig.~\ref{fig:conversation}. MarineGPT could recognize the marine objects present in marine images and generate corresponding responses aligned with the user intent. The generated responses cover the detailed information, addressing the user query. Through a user-friendly interface, MarineGPT could enable individuals to contribute data and increase public awareness about marine biodiversity and its significance.

\begin{figure}[t]
\begin{center}
\includegraphics[width=\textwidth]{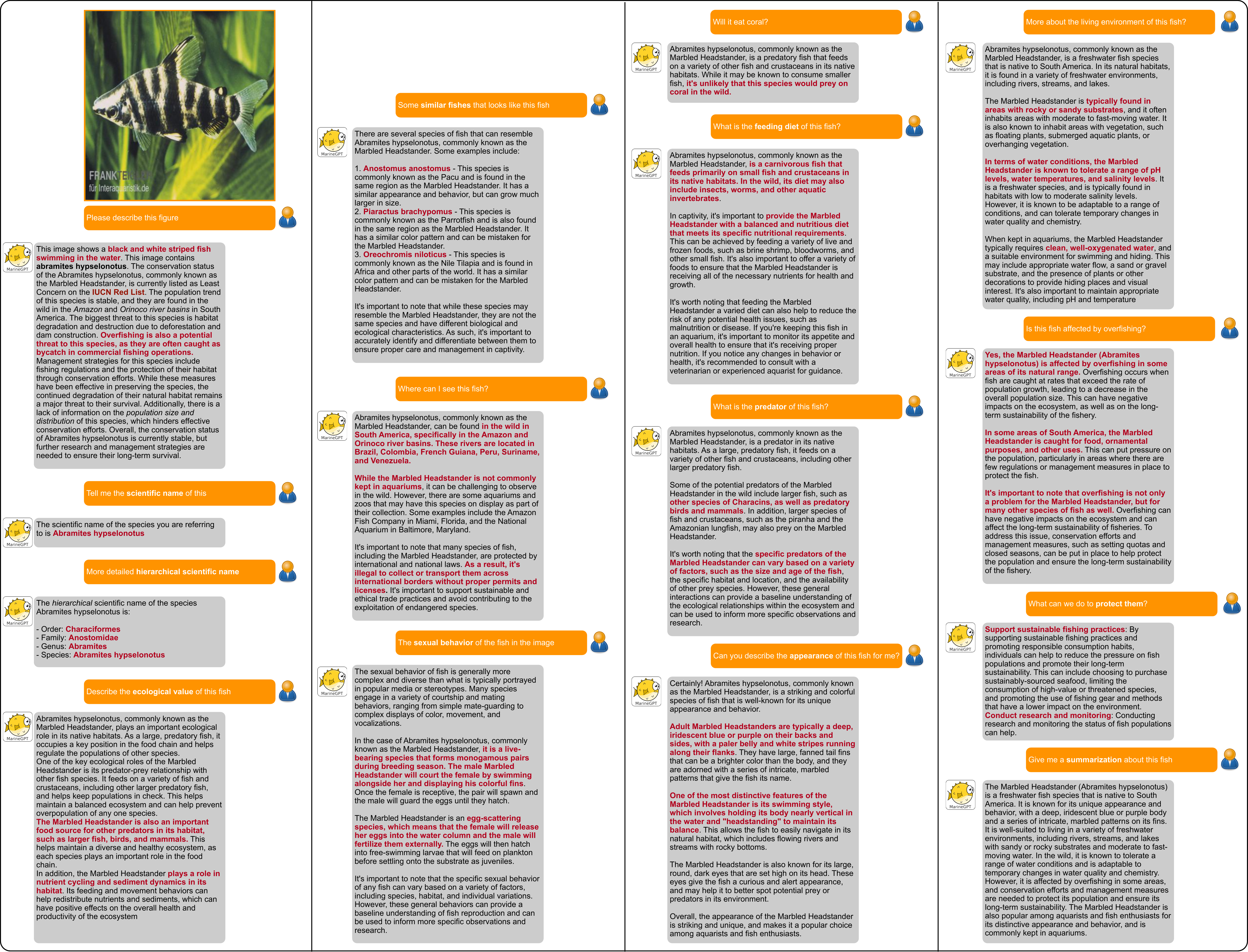}
\end{center}
\caption{MarineGPT could yield comprehensive marine and biological knowledge delivered to the users so that the users could obtain a full understanding of the recognized marine organism.}
\label{fig:comprehensive}
\end{figure}

\noindent\textbf{Comprehensive analysis}. One specific feature of our MarineGPT is that users can comprehensively understand the recognized marine object. We provide the illustration by choosing one fish image with ``Abramites hypselonotus'' as demonstrated in Fig.~\ref{fig:comprehensive}. In the multi-round conversations, the users could ask any question they are interested in the recognized marine species. MarineGPT could generate sensitive, domain-specific, and scientific responses for the users. Through training based on the instruction-following multi-modal data, our model could understand the user intent and generate required correspondences.

\begin{figure}[t]
\begin{center}
\includegraphics[width=\textwidth]{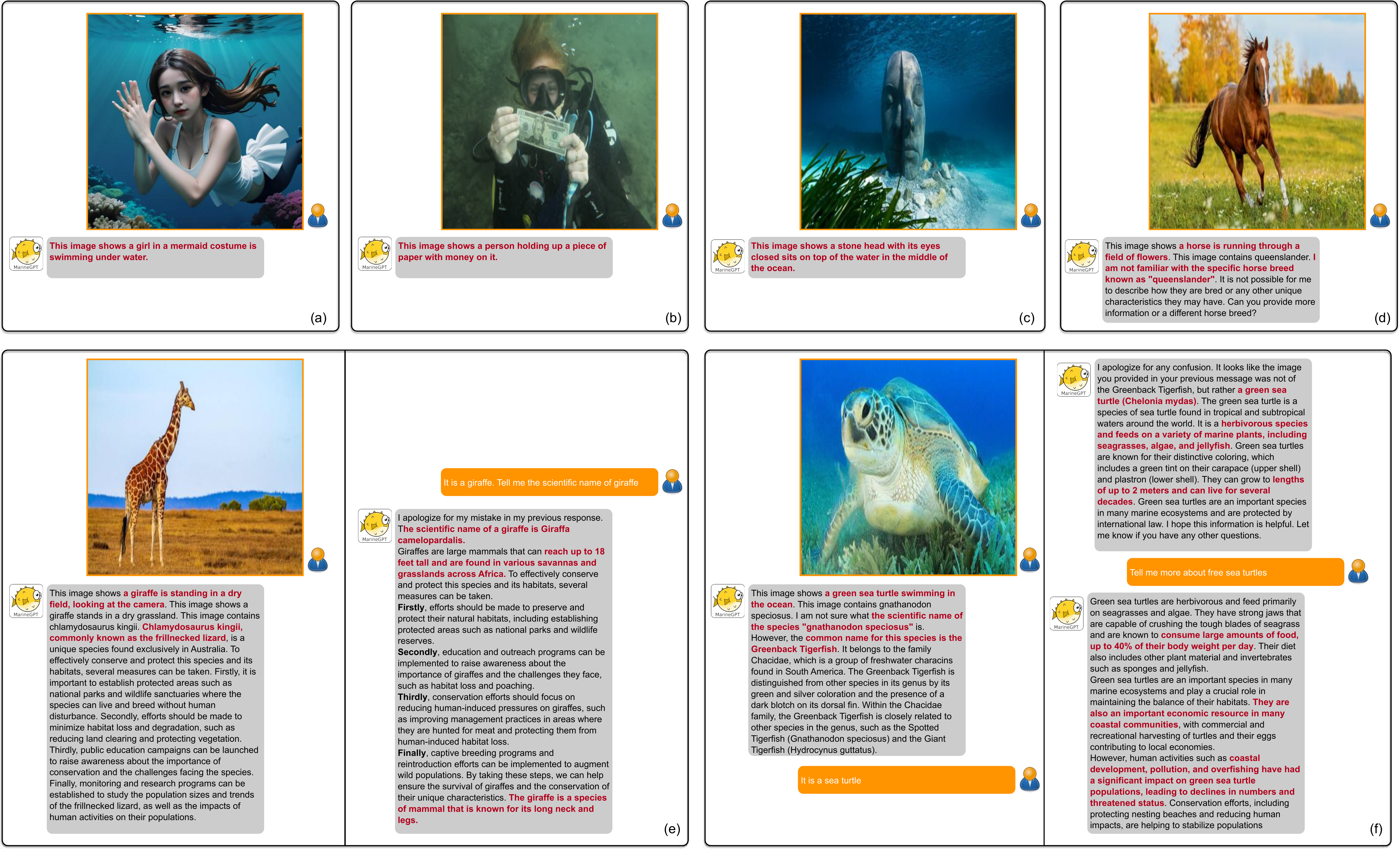}
\end{center}
\caption{The failure cases of proposed MarineGPT.}
\label{fig:failure}
\end{figure}

\textbf{Failure cases}. There are still some failure cases in our MarineGPT. It will refrain from generating long and informative responses for images that contain instances, that are not strictly ``marine creatures'' as reported in Fig.~\ref{fig:failure} a), b) and c). We have also explored whether MarineGPT suffers from knowledge forgetting in Fig.~\ref{fig:failure} d) and e). In d), MarineGPT could effectively recognize the horse and describe its behavior. But it will generate bizarre responses. As illustrated in Fig.~\ref{fig:failure} e), MarineGPT could recognize the instance accurately, but it misattributes the recognized organism as another species during the explanation procedure, leading to the in-context hallucination. With the feedback from the user, MarineGPT could provide the corresponding informative responses for the users. MarineGPT will also make similar mistakes for the marine images as demonstrated in Fig.~\ref{fig:failure} f). We attribute this failure to the inefficient ability of LLMs in the marine field and the lack of instruction-following data for such conceptions. Fine-tuning the LLMs to the marine domain with corresponding instruction-following data may effectively alleviate such issues.

\subsection{Specialized Features of MarineGPT}
Our MarineGPT has the following potential applications and broader impact on the marine field.
\begin{itemize}
    \item \textbf{Scale up marine organism recognition}. By integrating machine learning algorithms, MarineGPT could continually learn from user interactions, enhancing its species recognition accuracy and knowledge base over time. Scientists in marine biology and related fields can use the system for species identification, data retrieval, and ecological research. This accelerates the acquisition of biodiversity data, supporting research and decision-making processes.  
    \item  \textbf{Monitoring}. Our MarineGPT could be effectively integrated for accessing the marine species diversity, abundance data, and community structure. Furthermore, we could also perform the long-term marine species morning, which provides a unique opportunity to quantitatively examine changes in benthic habitat over time. It not only aids in marine object identification but also offers valuable insights into marine organisms' conservation status and ecological importance. This information could be used to prioritize conservation actions and monitor endangered or threatened species.  
    \item \textbf{Centralized platform}. Our MarineGPT provides a centralized platform for marine researchers, biologists, and institutions to share data, enable collaboration on marine research, and contribute to a global repository of marine species information. As the system recognizes and answers questions in a nearly real-time manner, it would facilitate on-the-spot data collection and recognition during field research or citizen science efforts. In this way, we can enable rapid and large-scale data collection, surpassing the limitations of manual expert identification efforts. Such a large-scale marine database is crucial for understanding species interactions, feeding diets, and ecosystem dynamics, leading to better conservation and management strategies for marine ecosystems.
    \item \textbf{Interdisciplinary research}. The system serves as an educational tool, fostering greater understanding and appreciation for marine life among users. It encourages citizen science participation and promotes environmental awareness and stewardship. Our method and MarineGPT empower users to access information on endangered or threatened species, aiding conservationists and policymakers in implementing targeted conservation measures to protect marine biodiversity. MarineGPT could also be integrated into educational tools, enhancing marine biology education and fostering a deeper appreciation for marine life among students and the public. Students, educators, and the general public could engage with the system, fostering a deeper understanding and appreciation for marine life and conservation. Collaborative efforts would improve the quality and accessibility of marine object information globally. 
    \item \textbf{General public access}. Researchers and enthusiasts could quickly access species information without requiring manual identification or extensive literature searches. Conservationists can access information on endangered species and use the system's data to make informed decisions for marine conservation efforts. Teachers and students in marine biology or environmental studies can utilize the system as an educational tool to enhance learning about marine species and ecosystems. Enthusiastic individuals interested in marine life can actively contribute to biodiversity monitoring and research by using the system during their marine observations or expeditions. Scuba divers, snorkelers, and ocean enthusiasts can utilize the system to identify marine species encountered during their underwater adventures. 
\end{itemize}

\section{Discussions}
\noindent\textbf{Hallucination}. Our MarineGPT inherits the hallucination problem from the frozen LLMs. Future advancements in more powerful LLMs are expected to alleviate this issue. Moreover, it is highly worthwhile to explore appropriate approaches to promote the ability of LLMs for domain-specific understanding. Through this, we could generate more domain-specific descriptions and knowledge for the images. We leave this as our future work.

\noindent\textbf{Video-centric MarineGPT}. The marine videos have not been considered integrated into chatting yet. The reasons probably come from the difficulty of accurately understanding non-static visual scenes. Besides, mitigating the modality gap between video and text~\citep{li2023videochat,zhang2023video}, which typically requires handling the temporary information while simultaneously considering both visual signals and audio signals. It is more challenging than bridging the gap between image and text. Our next work is to analyze the video with temporary collections, on how to summarize the relationships between different marine organisms. The massive video-caption pairs, as well as higher-quality and domain-specific visual-instruction-tuning datasets, are emergently required.

\section{Conclusion}
In this paper, we propose the first marine-specific multi-modal large language model, MarineGPT, which could generate more sensitive, informative, and scientific responses than existing general-purpose MLLMs. We present the largest marine image-text dataset with 5 million pairs and instruction-following cross-modality vision-language data for adapting a general-purpose AI assistant to a marine expert model. We propose a standard and valuable pipeline to develop a domain-specific expert model that involves specific knowledge. We will release our trained models and designed instructions to the research community to foster future research in both academic and industrial communities.

\bibliography{ref}
\bibliographystyle{iclr2024_conference}
\newpage
\end{document}